\documentclass[10pt,twocolumn,letterpaper]{article}

\usepackage{iccv}
\usepackage{times}
\usepackage{epsfig}
\usepackage{graphicx}
\usepackage{amsmath}
\usepackage{amssymb}
\usepackage{subcaption}
\usepackage{booktabs}
\usepackage{tabularx}
\usepackage[accsupp]{axessibility}
\newcommand{\mat}{}

\usepackage[pagebackref=true,breaklinks=true,letterpaper=true,colorlinks,bookmarks=false]{hyperref}
\usepackage[dvipsnames]{xcolor}

\iccvfinalcopy 

\def\iccvPaperID{****} 

\ificcvfinal\fi

\begin{document}

\title{Better Aggregation in Test-Time Augmentation}

\author{Divya Shanmugam\\
MIT CSAIL\\
{\tt\small divyas@mit.edu}
\and
Davis Blalock\\
MIT CSAIL\\
{\tt\small dblalock@mit.edu}
\and 
Guha Balakrishnan \\
Rice University \\
{\tt\small guha@rice.edu}
\and 
John Guttag \\
MIT CSAIL \\
{\tt\small guttag@csail.mit.edu}
}

\maketitle
\ificcvfinal\thispagestyle{empty}\fi

\begin{abstract}
Test-time augmentation---the aggregation of predictions across transformed versions of a test input---is a common practice in image classification. Traditionally, predictions are combined using a simple average. In this paper, we present 1) experimental analyses that shed light on cases in which the simple average is suboptimal and 2) a method to address these shortcomings. A key finding is that even when test-time augmentation produces a net improvement in accuracy, it can change many correct predictions into incorrect predictions. We delve into when and why test-time augmentation changes a prediction from being correct to incorrect and vice versa. Building on these insights, we present a learning-based method for aggregating test-time augmentations. Experiments across a diverse set of models, datasets, and augmentations show that our method delivers consistent improvements over existing approaches.
\end{abstract}


\section{Introduction} 
Data augmentation---the expansion of a dataset by adding transformed copies of each example---is a common practice in image classification. Typically, data augmentation is performed when a model is being trained. However, it can also be used at test-time to obtain greater robustness \cite{prakash2018deflecting, song2017pixeldefend, cohen2019certified}, improved accuracy \cite{krizhevsky2012imagenet, szegedy2015going, simonyan2014very, jin2018deep, matsunaga2017image}, or estimates of uncertainty \cite{matsunaga2017image, smith2018understanding, ayhan2018test, wang2019aleatoric}.
Test-Time Augmentation (TTA) entails pooling predictions from several transformed versions of a given test input to obtain a ``smoothed'' prediction. For example, one could average the predictions from various cropped versions of a test image, so that the final prediction is robust to any single unfavorable crop.

\begin{figure}[t!]
\begin{center}
   \includegraphics[width=1\linewidth]{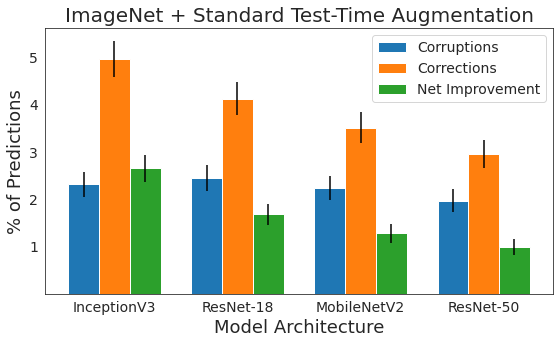}
\end{center}
\vspace{-.5cm}
   \caption{\textbf{Percentage of predictions corrected (orange) and corrupted (blue) by standard TTA}. Past work on TTA typically only examines the net improvement (green). This paper analyzes how standard TTA, which simply averages model predictions on transformed versions of a test image, can produce corruptions, and proposes a method that accounts for these factors.}
\label{fig:teaser}
\end{figure}

TTA is popular because it is easy to use. It is simple to put into practice with off-the-shelf libraries \cite{pytorch, keras}, makes no change to the underlying model, and requires no additional data. However, despite its popularity, there is relatively little research on the design choices involved in TTA. TTA depends on two choices: which augmentations to include, and how to aggregate the resulting predictions. We focus on the latter.  


Fig.~\ref{fig:teaser} shows the performance of a TTA policy that includes flips, crops, and scales applied to several models on ImageNet \cite{deng2009imagenet}. While the net improvement (green) is positive for each network architecture, a sizeable number of predictions are also changed to be incorrect (blue). Past TTA work typically only examines the net improvement, without considering why a TTA policy may actually degrade performance for many classes.

The goal of our work is twofold: (1) to understand which predictions TTA changes and why for a particular model and dataset and (2) to develop a method based on these insights that increases TTA performance. To do this, we first provide an empirical analysis of the corruptions introduced by TTA, and discuss implications for the design of TTA policies. Following this analysis, we present a learning-based method for TTA that depends upon these factors. In contrast to work on learning the choice of augmentations \cite{molchanov2020greedy, kim2020learning, sato2015apac}, we focus specifically on how we can learn to \emph{aggregate} augmentation predictions. The  solution we propose—learning optimal weights per augmentation, for a given dataset and model—can be applied in conjunction with other methods.




The proposed method represents a lightweight replacement for the simple average. Our method can offer a Top-1 accuracy increase of up to 2.5\%, and is nearly free in terms of model size, training time, and implementation burden. Our contributions are as follows:
\begin{itemize}
    \itemsep0em
    \item We provide insights into TTA that reveal why certain predictions are changed from correct to incorrect, and vice versa. We derive these insights from extensive experiments on ImageNet and Flowers-102 and include practical takeaways for the use of TTA. 

    \item We develop a TTA aggregation method that learns to aggregate predictions from different transformations for a given model and dataset. Our method significantly outperforms existing approaches, providing consistent accuracy gains across numerous architectures, datasets, and augmentation policies. We also show that the combination of TTA with smaller models can match the performance of larger models. 
\end{itemize}


\section{Related Work}

Image augmentation at test-time has been used to measure model uncertainty \cite{matsunaga2017image, smith2018understanding, ayhan2018test, wang2019aleatoric, bahat2020classification}, to attack models \cite{su2019one, moosavi2016deepfool, goodfellow2014explaining}, to defend models \cite{prakash2018deflecting, song2017pixeldefend, cohen2019certified}, and to increase test accuracy \cite{howard2013some, sato2015apac, he2016deep, simonyan2014very, szegedy2015going, krizhevsky2012imagenet}. 
 Because our focus is on test-time augmentation for the purpose of increasing image classification accuracy, we limit our discussion to work considering this problem.

Most works describing a test-time augmentation method for increasing classification accuracy present it as a supplemental detail, with a different methodological contribution being the focus of the paper. Krizhevsky \emph{et al.} \cite{krizhevsky2012imagenet} make predictions by ``extracting five 224 $\times$ 224 patches...as well as their horizontal reflections...and averaging the predictions made by the network’s softmax layer on the ten patches.'' He \emph{et al.} \cite{he2016deep} describe a similar setup and include an additional variation that incorporates rescaling of the input in addition to cropping and flipping. The cropping, scaling, and flipping combination is also employed by Simonyan \emph{et al.} \cite{simonyan2014very} and Szegedy \emph{et al.} \cite{szegedy2015going}, with differing details in each case. While most of these papers report results with and without test-time augmentation, none offers a systematic investigation into the merits of each augmentation function or how their benefits might generalize to other networks or datasets.

The works most closely related to our own are those of Sato \emph{et al.} \cite{sato2015apac}, Howard \emph{et al.} \cite{howard2013some}, Molchanov \emph{et al.} \cite{molchanov2020greedy}, and Kim \emph{et al.} \cite{kim2020learning}. The first seeks to improve classification accuracy by employing test-time augmentation. Their method samples augmentation functions randomly for each input, and makes predictions by averaging the log class probabilities derived from each transformed image. In contrast, we optimize the function that aggregates the predictions from each. Howard \emph{et al.} \cite{howard2013some} consider the problem of selecting a set of useful augmentations and proposes a method of choosing augmentations described as a ``greedy algorithm'' that ``starts with the best prediction and at each step adds another prediction until there is no additional improvement.'' The method is evaluated  on a single network and dataset, and does not learn to aggregate predictions as we do. Most recently, Molchanov \emph{et al.} \cite{molchanov2020greedy} propose Greedy Policy Search, which constructs a test-time augmentation policy by greedily selecting augmentations to include in a fixed-length policy. The predictions generated from the policy are aggregated using a simple average. Similarly, Kim \emph{et al.} \cite{kim2020learning} present a method to learn an instance-aware test-time augmentation policy. The method selects test-time augmentations with the lowest predicted loss for a given image, where the predicted loss is learned from the training data. 

Our work differs in that we focus on the factors that influence test-time augmentation and, given those factors, how we can learn to \emph{aggregate} augmentation predictions.


\section{Why weight augmentations differently?}

Typically, test-time augmentation methods aggregate model predictions by averaging \cite{krizhevsky2012imagenet, sato2015apac, kim2020learning}. While this is a reasonable approach, there are cases in which non-uniform weights are preferable. We analyze the errors simple averaging introduces on Flowers-102 and ImageNet to understand when non-uniform weights would be useful. 
\subsection{Setup}

\paragraph{Datasets} We use two datasets for our analysis: ImageNet (1000 classes) and Flowers-102 (102 classes). Our preprocessing pipeline is identical for ImageNet and Flowers-102: we resize the shortest dimension of each image to 256 pixels, followed by center cropping to produce a 256x256 image. We chose these datasets for their differences in difficulty and domain---the architectures we considered achieve $>$90\% accuracy on Flowers102 and 70-80\% on ImageNet. For each dataset, we apply normalization parameters based on the training set to each test image.


\paragraph{Models} We evaluate the performance of four architectures on ImageNet and Flowers-102: ResNet-18 \cite{he2016deep}, ResNet-50 \cite{he2016deep}, MobileNetV2 \cite{sandler2018mobilenetv2}, InceptionV3 \cite{szegedy2016rethinking}. We downloaded pretrained models from the PyTorch model zoo trained with an augmentation policy of horizontal flips and random crops \cite{pytorchModels}. To produce pretrained models for Flowers102, we use the finetuning procedure presented by \cite{cnn_finetune}. This procedure starts with a pretrained ImageNet network and freezes the weights in all but the last layer. The network is then trained on the new dataset for 100 epochs, using a batch size of 32, SGD optimizer (learning rate=.01, momentum=.9), and a dropout probability of .2. 

\paragraph{Augmentation Policies} We consider two augmentation policies. \emph{Standard}  reflects the typical augmentations used for TTA (flips, crops, and scales) and \emph{Expanded} includes a more comprehensive set of augmentations, such as intensity transforms. Readers interested in the specific augmentations may refer to the appendix. Each policy replaces the model's original predictions with an average of predictions on transformed images.

\indent The \emph{Standard} test-time augmentation policy produces 30 transformed versions per test image (a cross product of 2 flips, 5 crops, and 3 scales). The 5 crops correspond to the center crop and a crop from each corner. The three scale parameters are 1 (original image), 1.04 (4\% zoomed in) and 1.10 (10\% zoomed in), based on work that shows multi-scale evaluation improves model performance \cite{simonyan2014very}.

\indent The \emph{Expanded} test-time augmentation policy produces 128 transformations for each test image, consisting of 8 binary transforms from the PIL library \cite{pytorch} and 12 continuous transforms. We include 10 evenly-spaced magnitudes of each continuous transformation. We base this set of augmentations on AutoAugment \cite{cubuk2019autoaugment} with two major distinctions: 1) We make each augmentation function deterministic, to allow us to understand the specific relationship between an augmentation and model predictions, and 2) we do not consider combinations of these base transformations, because enumerating trillions of combinations would be infeasible.

\subsection{Overall results}


\begin{figure}
\centering
\begin{subfigure}[b]{\linewidth}
   \includegraphics[width=1\linewidth]{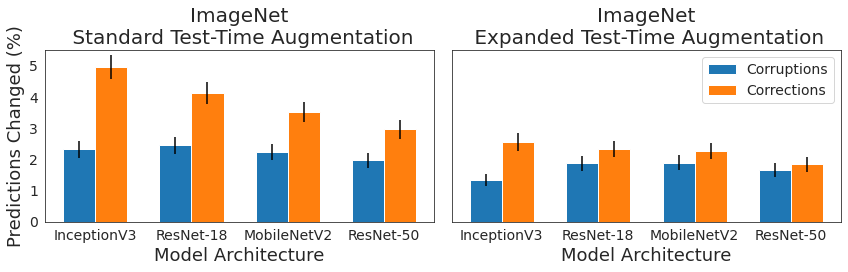}
\end{subfigure}

\begin{subfigure}[b]{\linewidth}
   \includegraphics[width=1\linewidth]{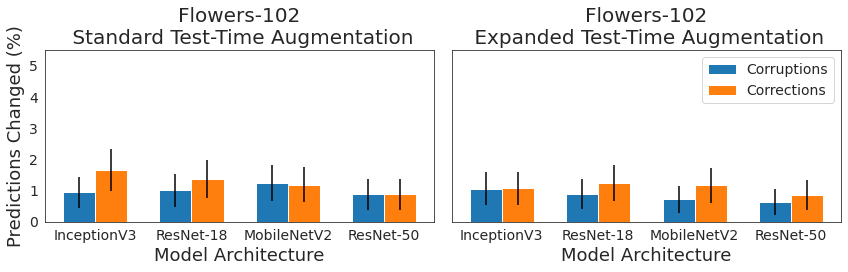}
\end{subfigure}
\caption{\textbf{Percentage of predictions corrected (orange) and corrupted (blue) by two TTA policies (Standard, Expanded).} Results for two datasets (ImageNet, Flowers-102) and four popular neural network models. Models are ordered by accuracy on classification task. We provide analysis of factors responsible for corruptions in Section 3.3.}
\label{tta_effect}
\vspace{-.7cm}
\end{figure}




Figure \ref{tta_effect} plots the percentages of corruptions and corrections introduced by the standard and expanded TTA policies on ImageNet and Flowers-102. The net effect of TTA is nearly always positive. However, the number of incorrect predictions introduced by the method represents a significant percentage of the changes introduced. In the context of ImageNet and ResNet-18, a little over one third of the labels changed by the standard TTA policy are incorrect.



Figure \ref{tta_effect} demonstrates that while one can expect a consistent improvement in accuracy from TTA, the magnitude of this improvement varies. We take a closer look at these results in the next sections to understand why TTA changes predictions to be correct to incorrect. 

\subsection{Biased Augmentation Sets}

\begin{figure}[htb!]
    \centering
    \begin{subfigure}{\linewidth}
   \includegraphics[width=1\linewidth]{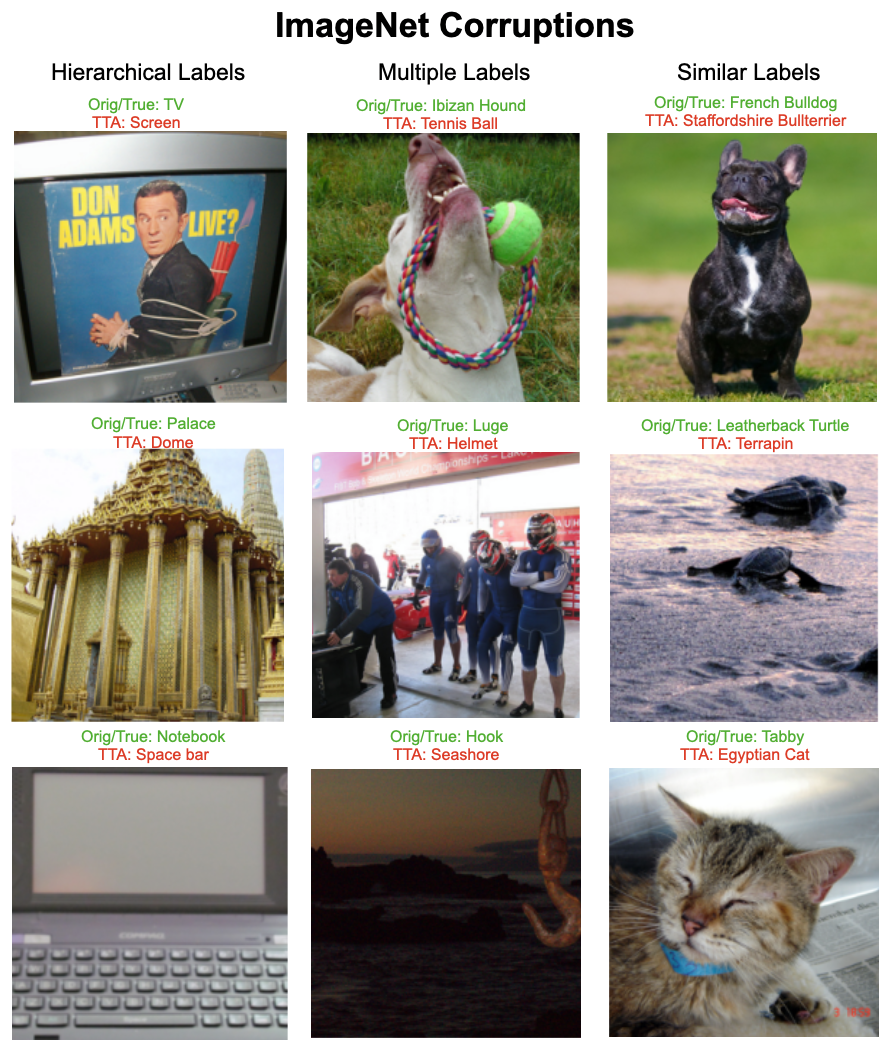}
\end{subfigure}

\begin{subfigure}{\linewidth}
   \includegraphics[width=1\linewidth]{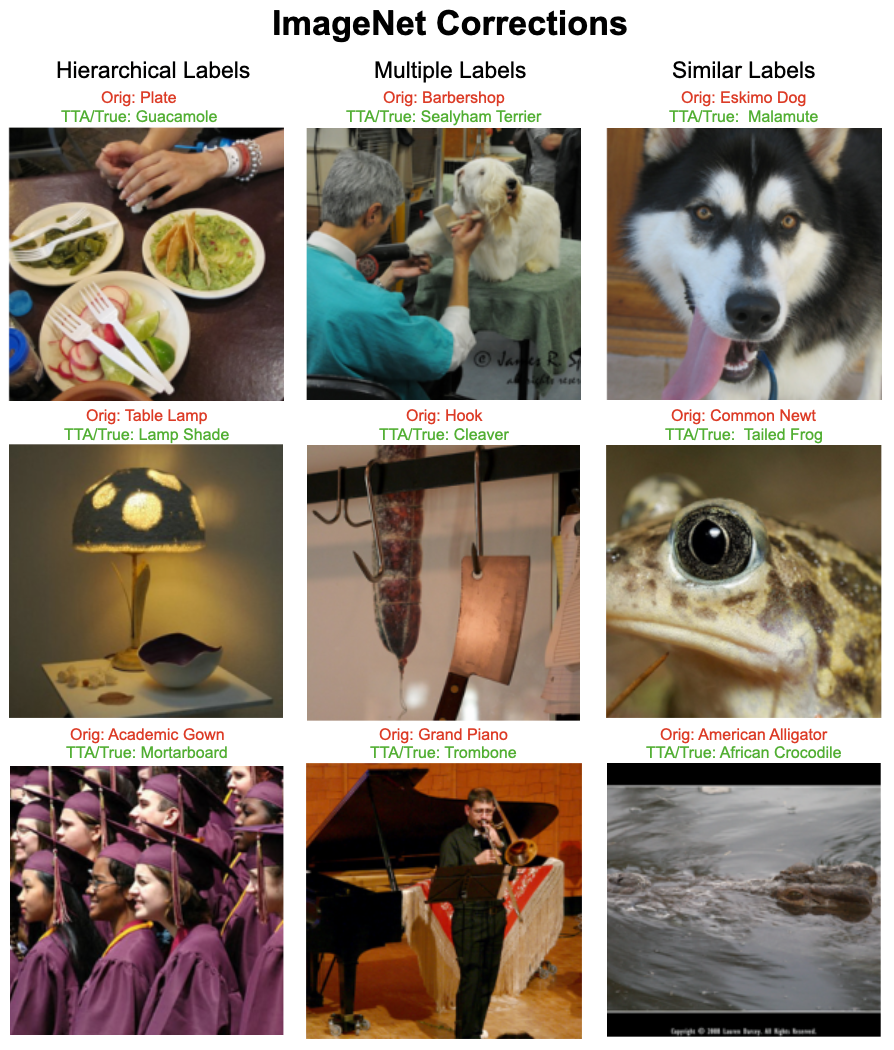}
\end{subfigure}
\caption{\textbf{TTA changes can be grouped into three types: hierarchical labels, multiple labels, and similar labels.} We include three examples from each type. TTA favors smaller and uncentered labels.}
   \label{tta_imnet_changes}
\end{figure}

Averaging implicitly assumes that the augmentation set has no influence on which predictions are corrected and which are corrupted. Examining TTA's effect on ImageNet, we show that this is not the case. In particular, crops introduce an inductive bias tied to the labeling scheme of the dataset. Instances that demonstrate this bias can be broken into three categories: hierarchical labels, multiple classes, and similar labels (Figure \ref{tta_imnet_changes}).


\emph{Hierarchical labels} include examples like (``plate", ``guacamole") and (``table lamp", ``lamp shade"). TTA often biases a prediction in favor of the smaller or uncentered component because of the crops included in the policy. Whether TTA produces a corruption or a correction depends on the assigned label. For example, Figure \ref{tta_imnet_changes} depicts an image where when the true label is ``palace" and TTA predicts ``dome." 

Other changed predictions correspond to images that contain objects from \emph{multiple classes} such as (``hook'',``cleaver'') and (``piano'', ``trombone'') (Figure \ref{tta_imnet_changes}). Recent work has noted this trait in ImageNet labels \cite{beyer2020we}. TTA produces incorrect labels by focusing on a different part of the image. Again, TTA predictions favor smaller objects because of crops. 

The last subset of major changes corresponds to confusing images, a product of \emph{similar labels} in the dataset (e.g., dog breeds). This subset is largely comprised of animals that are easily mistaken for one another. Crops and scales often increase confusion between classes when the resulting image emphasizes a non-distinguishing feature. For example, consider the ``Leatherback Turtle" image in Figure \ref{tta_imnet_changes}. One way in which Leatherback Turtles differ from terrapins is scale. As a result, the inclusion of a scaling augmentation naturally confuses the two. 






We see that the inductive bias arises because of a specific relationship between the augmentation set and the label space. Learning weights per augmentation allows us to identify and downweight augmentations that introduce inductive bias. We see this in later experiments.  

In practice, when using average aggregation, one should ensure that the augmentations have minimal correlation with the label space to avoid errors on images with hierarchical or multiple labels. When designing TTA policies for classes that are similar to one another, we should limit the magnitude of the transformations and choose augmentations that further distinguish confusing classes. For example, a zoomed-in version of an ``Egyptian Cat" is mistaken for a ``Tabby" because of a focus on fur (Figure \ref{tta_imnet_changes}) and smaller scales avoid such a mistake. TTAs that benefit well-separated classes are likely different from those that benefit often-confused classes.


\subsection{Class-Dependent Invariances}

The averaging aggregation strategy implicitly depends upon the same augmentation policy performing well for all input images. This is not the case when there are class-dependent invariances, as we see in Flowers-102.

\begin{figure}
    \centering
    \includegraphics[width=\linewidth]{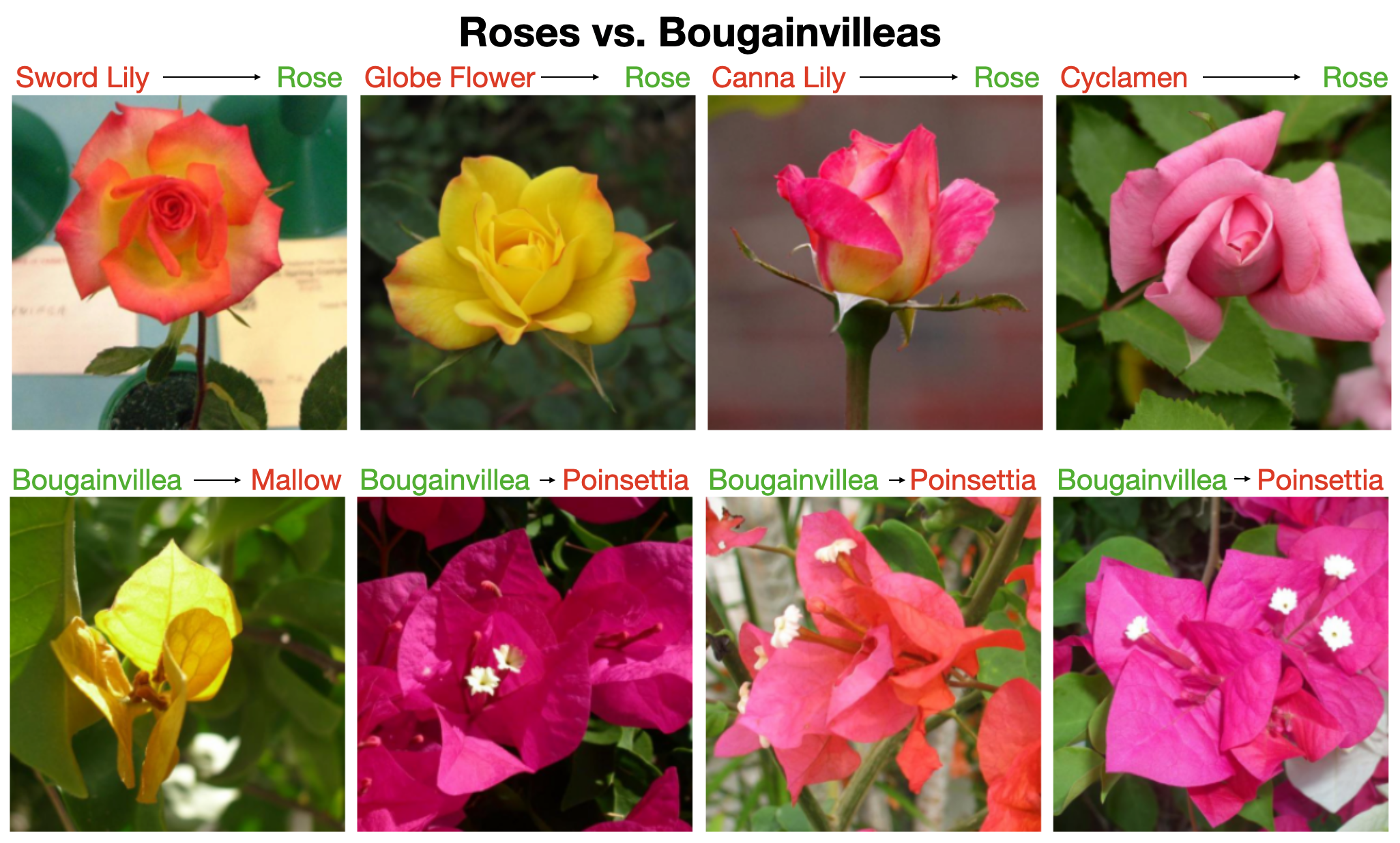}
    \caption{\textbf{Roses (top row) are most helped by TTA in Flowers-102, while Bougainvilleas (bottom row) are most harmed.} We show four cases of rose predictions being improved by TTA, and four cases where bougainvillea predictions are harmed. The white stamen of Bougainvilleas is both a distinguishing characteristic and prone to exclusion from certain crops, resulting in corruptions.}
    \label{fig:flowers102_best_worst}
    \vspace{-.5cm}
\end{figure}

Flowers-102 differs from ImageNet in many respects, such as dataset size, task difficulty, and class imbalance. Most importantly, it does not exhibit hierarchical labels or multiple labels. We show that crops have an intuitive effect on images from Flowers-102, similar to what we see on ImageNet. In particular, we show that crops can hurt flowers with smaller distinguishing features (see Figure \ref{fig:flowers102_best_worst}).


Consider images from the class most corrected by TTA (``Rose") and images from the class most corrupted (``Bougainvillea") in Figure \ref{fig:flowers102_best_worst}. The original predictions often mistake a rose for another flower with a similar color ( ``Globe Flower", ``Cyclamen") or shape (``Sword Lily", ``Canna Lily"). TTA may correct predictions for roses because crops maintain the petal texture, which differentiates roses from other classes. By including crops and zoomed-in portions of the image in the models' prediction, the model is better able to identify these textural differences.

The incorrect predictions introduced by TTA for ``Bougainvillea" are likely because of crops missing the cue of the white stamen, a distinguishing characteristic for the class. Moreover, crops may focus on a portion of the background (as with ``Mallow") and classify the image incorrectly. 

\begin{figure}[htb!]
  \centering
    \includegraphics[width=1\linewidth]{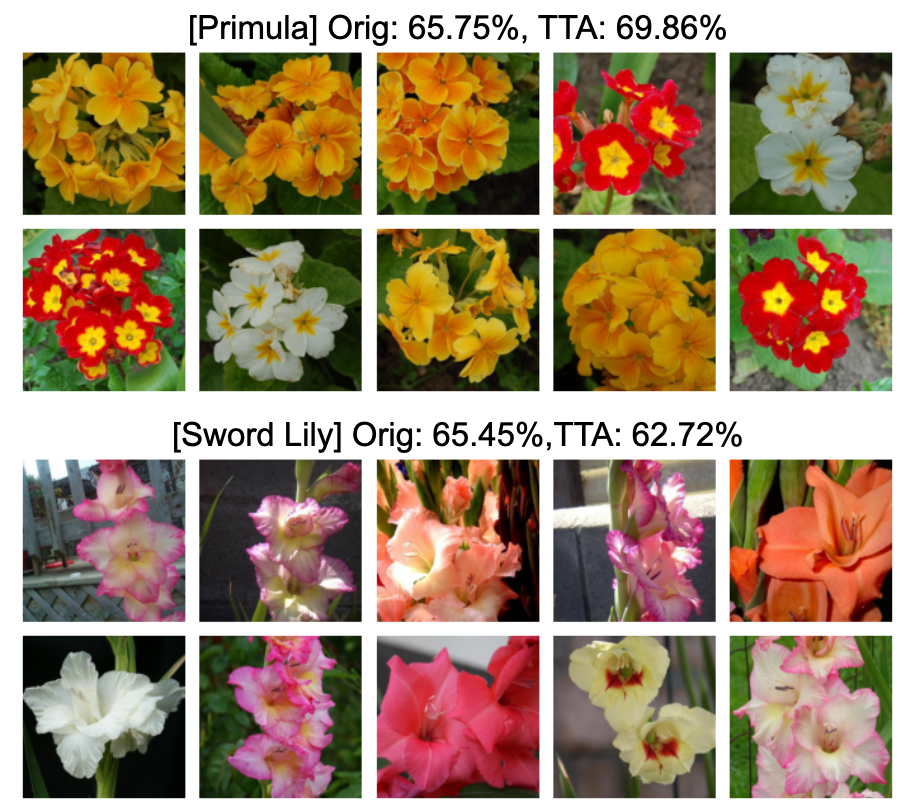}
    \caption{\textbf{Equally difficult classes produce different TTA behavior}. The training data for a class that TTA benefits (``Primula", top) look qualitatively different from a class TTA does not benefit (``Sword Lily", bottom).}
    \label{fig:flowers102_equally_difficult}
    \vspace{-.5cm}
\end{figure}

In Figure \ref{fig:flowers102_equally_difficult}, we compare images from two classes on which ResNet-50 performs equally well, ``Primula'' and ``Sword Lily.'' Interestingly, TTA improves performance on only one, ``Primula'' and not the other. ``Primula'' exhibits more consist texture, scale, and color than images of the ``Sword Lily.'' This observation suggests that the disparate effects of TTA could be caused by differences in variation within classes. Horizontal flips and random crops are not sufficient to account for the natural variation in ``Sword Lily" images, suggesting that this class would be better served with a non-uniform weighting TTA policy. In this case, augmentations would be downweighted for classes that do not exhibit the invariances TTA requires.

\section{Method}
In the previous section, we established cases in which weighting augmentations differently might address the errors introduced by TTA. We now present a simple learning model that learn these weights. We assume three inputs to our method:

\begin{enumerate}
\itemsep0em
\item A pretrained black-box classifier $f : \mathcal{X} \rightarrow \mathbb{R}^C$ that maps images to a vector of class probabilities. We use $\mathcal{X}$ to denote the space of images on which the classifier can operate and $C$ to denote the number of classes. We assume that $f$ is not fully invariant with respect to the augmentations.
\item A set of $M$ \textit{augmentation functions}, $\{a_m\}_{m=1}^{M}$. Each function $a_m: \mathcal{X} \rightarrow \mathcal{X}$ is a deterministic transform designed to preserve class-relevant information while modifying variables presumed to be class independent such as image scale or color balance. We use $A(x_i) \in \mathbb{R}^C$ to represent the matrix of $M$ augmentation predictions for input $x_i$.
\item A validation set of $N$ images $\mathbf{X} = \{x_i\}_{i=1}^N$ and associated labels $\{y_i\}_{i=1}^N$, $y_i \in \{1,\ldots,C\}$. We assume this set is representative of the test domain.
\end{enumerate}

Given these inputs, our task is to learn an \textit{aggregation function} $g: \mathbb{R}^{M \times C} \rightarrow \mathbb{R}^C$. Function $g$ takes a matrix of $C$ class logit predictions for $M$ augmented versions of a given image and uses them to produce one prediction in $\mathbb{R}^C$. We then apply a softmax layer to obtain a vector of class probabilities. Though $g$ can be arbitrarily complex, such as a multilayer neural network, we avoid adding significant size or latency. Therefore, we only consider functions of the form:

\begin{eqnarray}
    g(A(x_i)) &\triangleq& \sum_{m=1}^M (\mat{\Theta} \odot \mat{A(x_i)})_{m,*} 
\end{eqnarray}

\noindent where $\odot$ denotes an element-wise product and $\mat{\Theta} \in \mathbb{R}^{M \times C}$ is a matrix of trainable parameters. In words, $g$ learns a weight for each augmentation-class pair, and sums the weighted predictions over the augmentations to produce a final prediction. In scenarios where limited labeled training data is available, one may opt for $\mat{\Theta} \in \mathbb{R}^{M}$, where $\Theta$ has one weight for each augmentation:

\begin{eqnarray}
    g(A(x_i)) &\triangleq&  \mat{\Theta}^T \mat{A(x_i)}. 
\end{eqnarray}

We refer to (1) as \textit{Class-Weighted TTA}, or \textit{ClassTTA} and (2) as \textit{Augmentation-Weighted TTA}, or \textit{AugTTA}. We intend for $\Theta$ to represent an augmentation's importance to a final prediction and so impose a constraint that its elements must be nonnegative to favor interpretability of the resulting weights. We learn $\Theta$ by minimizing the cross-entropy loss between the true labels $y_i$ and the output of $g(A(x_i))$ using gradient descent. We choose between \textit{ClassTTA} and \textit{AugTTA} using a small held-out portion of the validation set and evaluate the performance of this method, in addition to the individual parameterizations.

\section{Experimental Evaluation}
 We evaluate the performance of our method across the datasets and architectures laid out in Section 3.1. We implemented our method in PyTorch \cite{pytorch} and employ an SGD optimizer with a learning rate of .01, momentum of .9, and weight decay of 1e-4. We apply projected gradient descent by clipping the weights to zero after each update to ensure the learned parameters are non-negative. We use the same optimization parameters across experiments and include them in the supplement. We train \textit{ClassTTA} and \textit{AugTTA} for 30 epochs, choose which to deploy on each dataset using a held-out validation set, and report our results on a held-out test set.

\paragraph{Datasets and Models}  We evaluate the performance of our method across the datasets and architectures laid out in Section 3.1. In addition, we evaluate on CIFAR-100 \cite{krizhevsky2009learning} and STL-10 \cite{coates2011analysis}. For each  dataset, we follow the preprocessing pipeline of \cite{pytorch-playground} and pad each image by 4 pixels to accommodate crops that maintain the original image size (32x32 and 96x96 respectively). We use popular pretrained networks for STL-10 and CIFAR-100 (a 5-layer CNN and a 7-layer CNN, respectively), courtesy of \cite{pytorch-playground}. 

\paragraph{Dataset Splits} We divide the released test sets into training (40\%), validation (10\%) and test (50\%) sets. We make training and validation sets available to methods that make use of labeled data. We make both the training and validation set available for methods that operate greedily,  so that each method makes use of the same amount of data.


\paragraph{Baselines}  We compare our method to four baselines:

\begin{itemize}
\itemsep0em
    \item \textit{Raw}: The original model's predictions, with no TTA.
    \item \textit{Mean}: Average logits across augmentations \cite{krizhevsky2012imagenet}. 
    \item \textit{Max}: Maximum logit across augmentations \cite{hendrycks2016baseline}.
    \item \textit{GPS}: Greedy Policy Search \cite{molchanov2020greedy}. GPS uses a parameter N, for the number of augmentations greedily included in a policy. We set this parameter to 3, in line with experiments reported in the original paper. GPS makes use of all labeled data (both the training and validation set).
\end{itemize}

While these baselines reflect existing work, they are not the only ways one could aggregate test-time augmentation predictions. Towards this end, we construct two other baselines: 1) learning to predict augmentation weights directly from an image and 2) learning to predict a mixture of \emph{Mean} and \emph{Raw} from an image. Our method dominates these constructed baselines in all experiments. We include these results in the supplement, in case they are useful to researchers pursuing similar ideas. 

\paragraph{Statistical Significance} We use a pairwise t-test to measure the statistical significance of our results and report standard deviations over 5 random subsamples of the test set.

\begin{table*}
\renewcommand{\arraystretch}{.75}
  \setlength\extrarowheight{-3pt}
\begin{tabularx}{\textwidth}{X|lllllll}
\toprule
 Dataset &    Model &               Original &               Max &              Mean &               GPS &              Ours \\
 \midrule
    Flowers102 &   MobileNetV2 &   $90.28 \pm 0.10$ &  $90.17 \pm 0.25$ &   $90.47 \pm 0.20$ &  $88.28 \pm 0.17$ &   $\mathbf{92.62 \pm 0.10}$ \\\\
    Flowers102 &   InceptionV3 &  $89.28 \pm 0.08$ &  $89.59 \pm 0.15$ &  $90.07 \pm 0.22$ &  $89.93 \pm 0.16$ &  $\mathbf{91.16 \pm 0.21}$ \\\\
    Flowers102 &     ResNet-18 &  $89.78 \pm 0.17$ &  $89.47 \pm 0.11$ &  $90.21 \pm 0.23$ &  $90.01 \pm 0.22$ &  $\mathbf{91.02 \pm 0.17}$ \\\\
    Flowers102 &     ResNet-50 &  $\mathbf{91.72 \pm 0.18}$ &  $91.61 \pm 0.08$ &  $\mathbf{91.96 \pm 0.27}$ &  $\mathbf{92.03 \pm 0.09}$ &  $\mathbf{92.02 \pm 0.16}$ \\\\
    ImageNet &   MobileNetV2 &  $71.38 \pm 0.06$ &   $72.50 \pm 0.13$ &  $\mathbf{72.69 \pm 0.06}$ &   $72.50 \pm 0.11$ &  $72.43 \pm 0.08$ \\\\
    ImageNet &   InceptionV3 &  $69.66 \pm 0.12$ &   $71.8 \pm 0.09$ &  $72.45 \pm 0.13$ &   $71.57 \pm 0.10$ &  $\mathbf{72.79 \pm 0.02}$ \\\\
    ImageNet &     ResNet-18 &   $69.37 \pm 0.1$ &  $70.26 \pm 0.13$ &  $\mathbf{71.02 \pm 0.13}$ &    $70.8 \pm 0.1$ &   $\mathbf{71.06 \pm 0.10}$ \\\\
    ImageNet &     ResNet-50 &  $75.78 \pm 0.08$ &  $76.62 \pm 0.08$ &  $\mathbf{76.91 \pm 0.09}$ &  $\mathbf{76.73 \pm 0.11}$ &  $\mathbf{76.75 \pm 0.14}$ \\\\
    CIFAR100 &  CNN-7 &  $74.15 \pm 0.18$ &  $75.00 \pm 0.31$ &  $75.48 \pm 0.11$ &  $75.45 \pm 0.21$ &  $\mathbf{75.92 \pm 0.20}$ \\\\
    STL10 &     CNN-5 &  $77.92 \pm 0.19$ &  $77.76 \pm 0.22$ &  $\mathbf{78.58 \pm 0.25}$ &  $\mathbf{78.32 \pm 0.17}$ &  $\mathbf{78.52 \pm 0.31}$ \\\\
    \toprule
    \end{tabularx}
\caption{TTA method performance (Top-1 Accuracy) given \emph{standard} augmentation policy.}
\label{standard_table}
\end{table*}

\begin{table*}
  \setlength\extrarowheight{-3pt}
\renewcommand{\arraystretch}{.75}
\begin{tabularx}{\textwidth}{X|llllllll}
\toprule
Dataset &    Model &               Original &               Max &              Mean &               GPS &              Ours \\
\midrule
Flowers102 &   MobileNetV2 &  $90.94 \pm 0.16$ &  $86.85 \pm 0.24$ &  $91.14 \pm 0.08$ &  $91.34 \pm 0.16$ &  $\mathbf{92.49 \pm 0.20}$ \\\\
Flowers102 &   InceptionV3 &  $89.17 \pm 0.33$ &  $87.89 \pm 0.20$ &  $89.20 \pm 0.23$ &  $89.43 \pm 0.16$ &  $\mathbf{91.02 \pm 0.26}$ \\\\
Flowers102 &     ResNet-18 &  $89.20 \pm 0.10$ &  $83.30 \pm 0.19$ &  $89.47 \pm 0.09$ &  $\mathbf{89.90 \pm 0.24}$ &  $\mathbf{89.78 \pm 0.16}$ \\\\
Flowers102 &     ResNet-50 &  $92.37 \pm 0.13$ &  $89.39 \pm 0.19$ &  $92.48 \pm 0.11$ &  $92.57 \pm 0.21$ &  $\mathbf{93.29 \pm 0.21}$ \\\\
ImageNet &   MobileNetV2 &  $71.18 \pm 0.05$ &  $67.65 \pm 0.08$ &  $71.84 \pm 0.12$ &  $\mathbf{72.49 \pm 0.09}$ &  $\mathbf{72.57 \pm 0.09}$ \\\\
ImageNet &   InceptionV3 &  $69.51 \pm 0.08$ &  $66.00 \pm 0.13$ &  $70.85 \pm 0.11$ &  $\mathbf{71.05 \pm 0.08}$ &  $\mathbf{71.02 \pm 0.06}$ \\\\
   ImageNet &     ResNet-18 &  $69.62 \pm 0.15$ &  $66.56 \pm 0.12$ &  $70.11 \pm 0.13$ &  $\mathbf{70.91 \pm 0.05}$ &  $\mathbf{70.89 \pm 0.04}$ \\\\
  ImageNet &     ResNet-50 &  $75.53 \pm 0.06$ &  $71.99 \pm 0.15$ &  $75.87 \pm 0.17$ &  $76.12 \pm 0.08$ &  $\mathbf{76.36 \pm 0.10}$ \\\\
CIFAR100 &  CNN-7 &  $74.37 \pm 0.12$ &  $63.90 \pm 0.22$ &  $73.41 \pm 0.13$ &  $\mathbf{75.07 \pm 0.32}$ &  $73.18 \pm 0.21$ \\\\

STL10 &     CNN-5 &  $78.04 \pm 0.18$ &  $74.77 \pm 0.12$ &  $\mathbf{79.02 \pm 0.21}$ &  $\mathbf{78.81 \pm 0.27}$ &  $\mathbf{79.27 \pm 0.22}$ \\\\
\toprule
\end{tabularx}
\caption{TTA method performance (Top-1 Accuracy) given \emph{expanded} augmentation policy.}
\label{expanded_table}
\vspace{-.5cm}
\end{table*}

\subsection{Standard TTA Policy}

\paragraph{Results} As shown in Table \ref{standard_table}, our method significantly outperforms all baselines (p-value=2e-7). Moreover, our method significantly outperforms the original model in all 8 comparisons (p-value=7e-10). Our method outperforms other baselines in 42 of the 50 individual trials summarized by Table \ref{standard_table}. Tables including results for Top-5 classification accuracy can be found in the supplement. 

On STL-10, our method, \emph{Mean}, and \emph{GPS} perform comparably. While our method learns to ignore augmentations that provide no additional information, the weighting for the remaining augmentations is roughly equivalent to the average. This helps identify which augmentations need not be included, thereby saving computation per image, but does not significantly improve performance. 

While \emph{Max} demonstrates predictive power in identifying out-of-distribution examples \cite{hendrycks2016baseline}, the same cannot be said for selecting which test-time augmentation lies closest to the training distribution. This is likely caused by the well-established phenomenon of miscalibration in neural networks \cite{guo2017calibration}.
\vspace{-.4cm}

\paragraph{Analysis} The method consistently chooses \emph{ClassTTA} on Flowers-102 and \emph{AugTTA} on ImageNet. This is likely because of the large number of classes in ImageNet (1000) and the relatively few examples per class (25) to learn from. For STL-10 and CIFAR-100, both \emph{AugTTA} and \emph{ClassTTA} converge to similar augmentation weightings, which suggests there are no significant class-dependent relationships with the standard TTAs.

Given enough data, \emph{ClassTTA} should provide a strict improvement over \emph{AugTTA}. Therefore, these results imply that \emph{ClassTTA} is best applied to datasets with few classes and sufficient labeled data. We include results for each parameterization in the appendix. In some cases, our method does worse than either individual parameterization -- this is because it makes use of a small hold-out validation set to decide between the two. This suggests that in some cases, it is more useful to select a parameterization based on domain knowledge and learn a more performant set of weights.

The benefit of simple averaging (\emph{Mean}) diminishes with larger networks. We also find that the magnitude of TTA-based  improvement is correlated with the number of examples per class (r=.95, p-value=.04). This suggests that the model relies on a large number of examples per class to identify invariances during training, so that TTA can exploit them during inference.

Our experiments also suggest that the combination of TTA with smaller networks can outperform larger networks without TTA and may be of use when deploying models in space-constrained settings. This can be seen in the higher performance of \emph{ClassTTA} applied to MobileNetV2 ($\sim$3.4 million parameters) compared to the original ResNet-50 model ($\sim$23 million parameters) on Flowers-102.


\subsection{Expanded TTA Policy}

\paragraph{Results} Table \ref{expanded_table} presents our results with a larger set of augmentations. Our method significantly outperforms the traditional averaging (p-value=2e-7). The results show that we outperform GPS (p-value=5e-7), exceeding its performance on 34 of the 50 trials. Once more, our method favors \textit{ClassTTA} for Flowers-102 and \textit{AugTTA} for ImageNet.  Results in the supplement show that \textit{ClassTTA} yields larger improvement for Flowers-102 and moderate improvements on ImageNet. \textit{ClassTTA} significantly outperforms the original model on all datasets (p-value=1e-6). In the case of MobileNetV2 and ImageNet, our method underperforms the best performing baseline (\emph{Mean}) because it selects \emph{ClassTTA} over \emph{AugTTA} using the validation set, when \emph{AugTTA} performs comparably to \emph{Mean}.
\paragraph{Analysis} Interestingly, many of the TTAs considered in this policy were not included in any model's train-time augmentation policy. Each model was trained with only two train-time augmentations: flips and crops. This suggests that useful test-time augmentations need not be included during training and may reflect dataset-specific invariances. 

The tradeoff in using an expanded set of TTAs is the increased cost at inference time. Each additional augmentation increases the batch size that must be passed through the network. This cost of an expanded set of augmentations may not be justified according to our results: the accuracy of \textit{ClassTTA} using a standard set of TTAs is comparable to accuracy of \textit{ClassTTA} using an expanded set of TTAs. This may be because the standard set of TTAs overlaps with the augmentations used during training. Further investigation is necessary to determine the relationship between train-time and test-time augmentation policies.


 
\subsection{Learned Weights}

The performance of \emph{AugTTA} and \emph{ClassTTA} demonstrate that there are cases where taking the mean of augmentation predictions is not optimal.

\begin{figure}[t]
  \centering
    \includegraphics[width=1\linewidth]{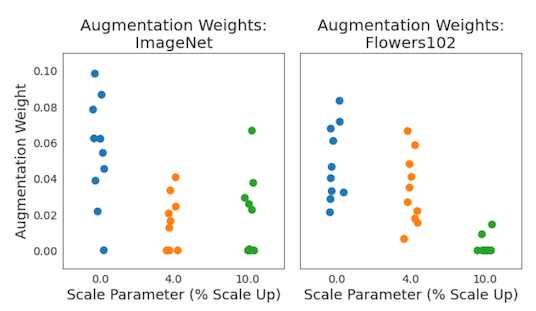}
\vspace{-.5cm}
    \caption{\textbf{Augmentations with higher scale parameters are weighted lower by our method.} Learned augmentation weights for each of the 30 augmentations included in the standard policy. Higher scales are weighted lower for both datasets.}
    \label{aug_weights}
 \vspace{-.6cm}
\end{figure}

Across all architectures on ImageNet, our method learns to exclude the augmentations that include a 10\% scale from the final prediction (Figure \ref{aug_weights}). This reflects our qualitative analysis suggesting that scales can introduce an undesirable inductive bias in the final predictions (Figure \ref{tta_imnet_changes}). 

While TTA performance given the expanded policy does not outperform TTA with the standard policy, the learned weights tell us more about a broader set of reasonable test-time augmentations for these datasets. For example, crop, translation, and blur augmentations are consistently weighted highly in the expanded policy setting. On the other hand, weights for contrasts, cut-outs, shearing, and brightness augmentations are consistently learned to be 0.

\begin{figure}[t]
\begin{center}
   \includegraphics[width=1\linewidth]{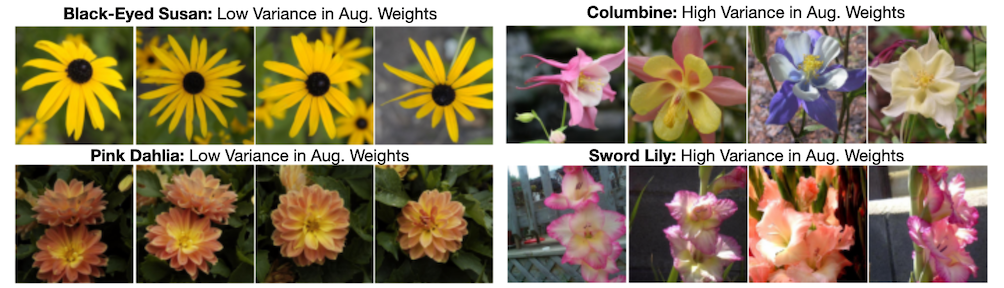}
\end{center}
\vspace{-.5cm}
   \caption{\textbf{Classes with higher variation in learned augmentation weights exhibit  higher input variation}. We show examples from two classes with the lowest (left) and  highest (right) variation in augmentation weights (using ResNet-50, Flowers-102, \emph{Standard} TTA policy).}
\label{class_specific_weights}
\vspace{-.5cm}
\end{figure}

Similarly, while \emph{ClassTTA} frequently underperforms \emph{AugTTA}, the class-specific weights offer insight into the training images for each class. Specifically, classes with higher variance in learned augmentation weights exhibit higher input variation (Figure \ref{class_specific_weights}).

Supporting plots for additional architectures and augmentation comparisons and the expanded test-time augmentation policy are included in the appendix. In each case, augmentations with higher scale parameters (corresponding to more zoomed-in images) are weighted lower.

\subsection{Computational Cost}

The benefit of TTA comes at the cost of repeated inference. The computational cost of TTA is offset by 1) potential for batched inference, thereby reducing inference time and 2) the ease-of-use compared other methods for improving model accuracy (e.g., model retraining). 
    
    Implemented naively, the cost scales linearly with the magnitude of the TTA policy. However, one can also use the per-augmentation weights to \emph{decide} which augmentations to generate. For ResNet-50 on ImageNet, AugTTA learns non-zero weights for only 37 of the 128 augmentations in \emph{Expanded} TTA policy (28\%). On Flowers-102, only 20
(16\%) have non-zero weights, demonstrating that one can also use this method to save computation.

\section{Discussion}

In this paper, we investigate when test-time augmentation works, and when it does not. Through an analysis of two widely-used datasets---ImageNet and Flowers-102---we show that the predictions changed by TTA reveal how weighting augmentations differently can be useful. We build on these insights to construct a method that accounts for these factors and show that it outperforms existing TTA approaches across 4 datasets and 6 models. Analysis of the learned weights highlights useful test-time augmentations that lie outside the standard policy of flips, crops, and scales. 

The insights shared in this study can improve the field's understanding of how TTA changes model decisions. This work opens promising areas for future work:

\begin{itemize}
    \itemsep0em
    \item \emph{Targeted train-time augmentation policies}: TTA exploits a model's lack of invariance to certain transforms. A model could instead learn this invariance, as recent work has shown \cite{benton2020learning}. The success of TTA could highlight when and where there is a greater need for train-time augmentation and can inform a set of class-specific transforms to include during training. 
    \item \emph{Learned augmentations}: Learning the weights for each augmentation is only one way to build on the insights presented here. One could instead learn a set of augmentations. Past work on TTA considers common augmentations but it is worth considering a broader class of augmentations.
\end{itemize}


{\small
\bibliographystyle{ieee_fullname}
\bibliography{main}
}

\end{document}